\documentclass{article} 

\usepackage[preprint]{nips_2018}

\usepackage{amsmath,amsfonts,bm}

\def\eqref#1{equation~\ref{#1}}

\def\1{\bm{1}}

\DeclareMathAlphabet{\mathsfit}{\encodingdefault}{\sfdefault}{m}{sl}
\SetMathAlphabet{\mathsfit}{bold}{\encodingdefault}{\sfdefault}{bx}{n}

\usepackage{hyperref}
\usepackage{url}
\usepackage{graphicx}
\usepackage{subcaption}

\usepackage{booktabs}
\usepackage[title]{appendix}
\usepackage[parfill]{parskip}
\usepackage{lmodern}
\usepackage{siunitx}
\usepackage{amssymb}
\usepackage{pifont}
\usepackage{etoolbox}
\usepackage{booktabs}
\usepackage{multirow}

\usepackage[utf8]{inputenc} 
\usepackage[T1]{fontenc}    
\usepackage{hyperref}       
\usepackage{url}            
\usepackage{booktabs}       
\usepackage{amsfonts}       
\usepackage{nicefrac}       
\usepackage{microtype}      

\title{DEFactor: Differentiable Edge Factorization-based Probabilistic Graph Generation}

\author{Rim Assouel \\
Mila, Université de Montréal, QC\\
Benevolent.AI, London, UK\\
\texttt{rim.assouel@umontreal.ca}\\
\And 
Mohamed Ahmed\\
Benevolent.AI, London, UK\\
\texttt{mohamed.ahmed@benevolent.ai}\\
\And
Marwin H Segler\\
Benevolent.AI, London, UK\\
\texttt{marwin.segler@benevolent.ai}\\
\And
Amir Saffari\\
Benevolent.AI, London, UK\\
\texttt{amir.saffari@benevolent.ai}\\
\And
Yoshua Bengio\thanks{CIFAR Senior Fellow}\\
Mila, Université de Montréal, QC\\
\texttt{yoshua.bengio@mila.quebec}
}

\begin{document}
\maketitle
\begin{abstract}
Generating novel molecules with optimal properties is a crucial step in many industries such as drug discovery. 
Recently, deep generative models have shown a promising way of performing de-novo molecular design. 
Although graph generative models are currently available they either have a graph size dependency in their number of parameters, limiting their use to only very small graphs or are formulated as a sequence of discrete actions needed to construct a graph, making the output graph non-differentiable w.r.t the model parameters, therefore preventing them to be used in scenarios such as conditional graph generation. 
In this work we propose a model for conditional graph generation that is computationally efficient and enables direct optimisation of the graph. 
We demonstrate favourable performance of our model on prototype-based molecular graph conditional generation tasks.
\end{abstract}

\section{Introduction}
We address the problem of learning probabilistic generative graph models for tasks such as the conditional generation of molecules with optimal properties. More precisely we focus on generating realistic molecular graphs, similar to a target molecule (the prototype).

The main challenge here stems from the discrete nature of graph representations for molecules; which prevents us from using global discriminators that assess generated samples and back-propagate their gradients to guide the optimisation of a generator. This becomes a bigger hindrance if we want to either optimise a property of a molecule (graph) or explore the vicinity of an input molecule (prototype) for conditional optimal generation, an approach that has proven successful in controlled image generation \citep{cond_AC,infogan}.

Several recent approaches aim to address this limitation by performing indirect optimisation~\citep{JTVAE, Lesko, Li_graph}. You et al.~\cite{Lesko} formulate the molecular graph optimisation task in a reinforcement learning setting, and optimise the loss with policy gradient~\cite{yu2016seqgan}. However policy gradient tends to suffer from high variance during training. Kang and Cho~\cite{Cho} suggest a reconstruction-based formulation which is directly applicable to discrete structures and does not require gradient estimation. However, it is limited by the number of samples available. Moreover, there is always a risk that the generator simply ignores the part of the latent code containing the property that we want to optimise. Finally, Jin et al.~\cite{JTVAE} apply Bayesian optimisation to optimise a proxy (the latent code) of the molecular graph, rather than the graph itself.

In contrast, Simonovsky and Komodakis~\cite{martin} and De Cao and Kipf~\cite{Molgan} have proposed decoding schemes that output graphs (adjacencies and node/edge feature tensors) in a single step, and so are able to perform direct optimisation on the probabilistic continuous approximation of a graph. However, both decoding schemes make use of fixed size MLP layers which restricts their use to very small graphs of a predefined maximum size.

\begin{figure}[t!]
    \centering
    \begin{subfigure}[b]{0.8\textwidth}
        \includegraphics[height=3.5cm]{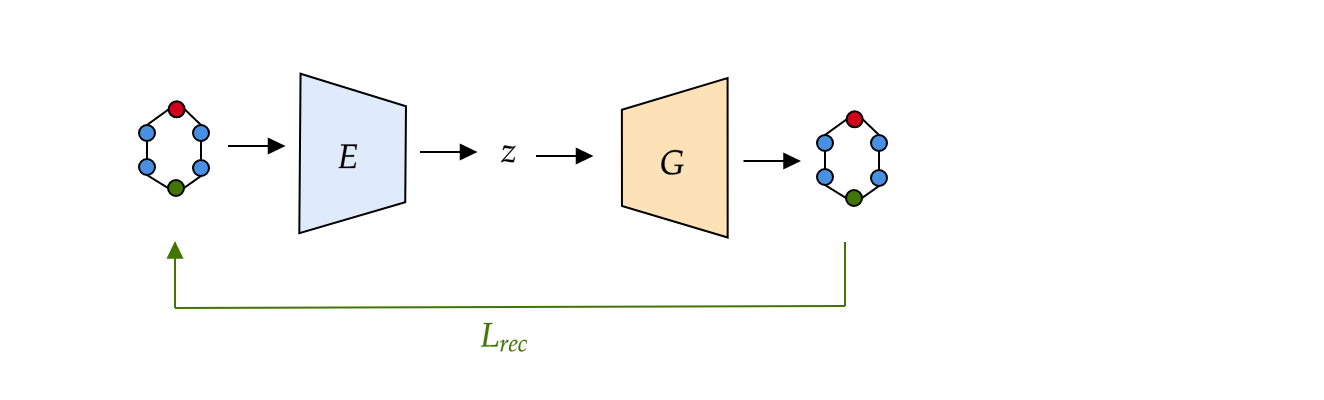}
        \caption{The full autoencoder (step \textbf{1} to \textbf{4})}
        \label{fig:sep1}
    \end{subfigure}
    ~ \vspace{-0.1cm}\\ 
    \begin{subfigure}[b]{0.8\textwidth}
        \includegraphics[height=3.3cm]{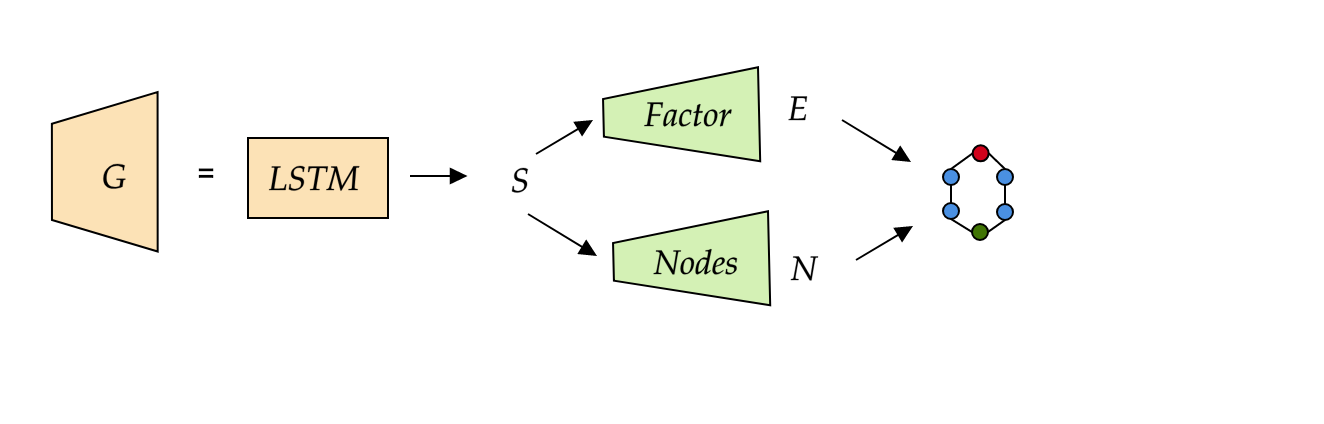}
        \caption{Expanding the steps (\textbf{3} for the LSTM and \textbf{4} for the factorization and node decoding) of the generator $G$ of the autoencoder}
        \label{fig:sep2}
    \end{subfigure}
    ~ \vspace{-0.1cm}\\ 
    \begin{subfigure}[b]{0.8\textwidth}
        \includegraphics[height=3.2cm]{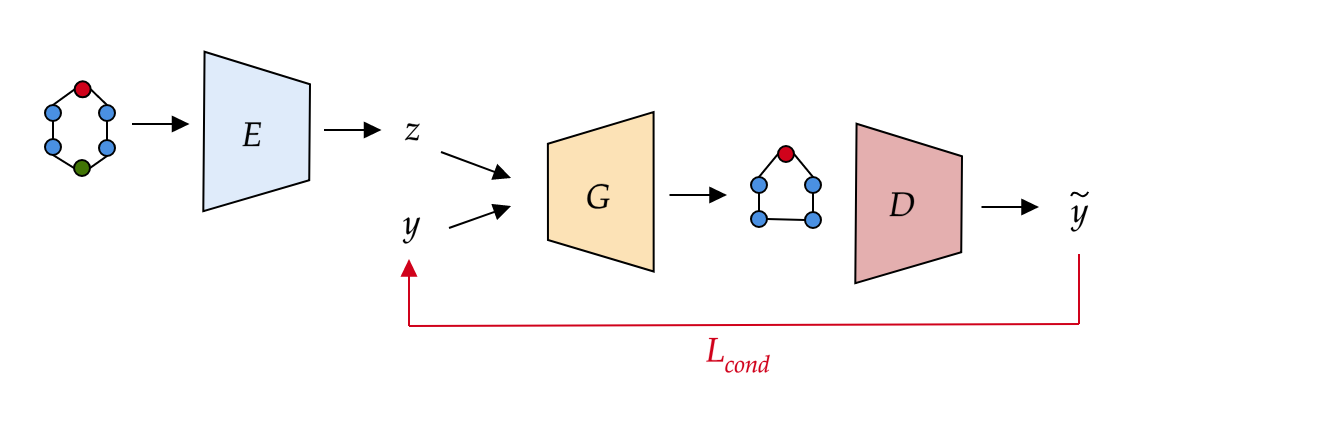}
        \caption{The conditional setting a discriminator $D$ that assesses the outputs and gives its feedback to the generator $G$. $L_{rec}$ (resp. $L_{cond}$) refers to the reconstruction (resp. conditional) loss described in section 3.3.}
        \label{fig:sep3}
    \end{subfigure}
    \caption{Overview of our molecule autoencoding  ((a) and (b)) and conditional generation (c) process. }\label{fig:full_model}
    
\end{figure}

Our approach (DEFactor) depicted in Figure~\ref{fig:full_model} aims to directly address these issues with a probabilistic graph decoding scheme that is end-to-end differentiable, computationally efficient w.r.t the number of parameters in the model and capable of generating arbitrary sized graphs (\autoref{sec:model}). We evaluate DEFactor on the task of constrained molecule property optimisation~\cite{JTVAE, Lesko} and demonstrate that our results are competitive with recent results. 



\section{Related work}

\paragraph{Lead-based Molecule Optimisation:}  The aim here is to obtain molecules that satisfy a target set of objectives, for example activity against a biological target while not being toxic \textit{or} maintaining certain properties, such as solubility. Currently a popular strategy is to fine-tune an pretrained generative model to produce/select molecules that satisfy a desired set of properties~\cite{marwin}. 

Bayesian optimisation is proposed to explore the learnt latent spaces for molecules in~\cite{gomez}, and is shown to be effective at exploiting feature rich latent representations~\cite{pmlr-v70-kusner17a, Dai2018SyntaxDirectedVA,  JTVAE}. In \cite{Li_deep, Li_graph} sequential graph decoding schemes whereby  conditioning properties can be added to the input are proposed. However these approaches are unable to perform direct optimisation for objectives. Finally~\cite{Lesko} reformulates the problem in a reinforcement learning setting, and objective optimisation is performed while keeping an efficient sequential-like generative scheme~\cite{GraphRNN}. 
 
\paragraph{Graph Generation Models:} Sequential methods to graph generation~\cite{GraphRNN,Li_graph,Lesko,Li_deep} aim to construct a graph by predicting a sequence of addition/edition actions of nodes/edges. Starting from a sub-graph (normally empty), at each time step a discrete transition is predicted and the sub-graph is updated. Although sequential approaches enable us to decouple the number of parameters in models from the the maximum size of the graph processed, due to the discretisation of the final outputs, the graph is still non-differentiable w.r.t. to the decoder's parameters.  This again prevents us from directly optimising for the objectives we are interested in.

In contrast to the sequential process~\cite{Molgan, martin} reconstruct probabilistic graphs. These methods however make use of fixed size MLP layers when decoding to predict the graph adjacency and node tensors.  This however limits their use to very small graphs of a pre-chosen maximum size.  They therefore restrict study and application to small molecular graphs; a maximum number of 9 heavy atoms, compared to approximately 40 in sequential models. 

We propose to tackle these drawbacks by designing a graph decoding scheme that is: 
\begin{itemize}
    \item \textbf{Efficient}: so that the number of parameters of the decoder does not depend on a fixed maximum graph size.
    \item \textbf{Differentiable}: in particular we would like the final graph to be differentiable w.r.t the decoder's parameters, so that we are able to directly optimise the graph for target objectives.
\end{itemize}



%
\section{DEFactor}\label{sec:model}
Molecules can be represented as graphs $G = (V, E)$ where atoms and bonds correspond to the nodes and edges respectively. Each node in V is labeled with its atom type which can be considered as part of its features. The adjacency tensor is given by  $E \in \left \{ 0,1 \right \}^{n \times n \times e}$ where $n$ is the number of nodes (atoms) in the graph and $e$ is the number of possible edge (bond) types. The node types are represented by a node feature tensor  $ N \in \left \{ 0,1 \right \}^{n \times d } $ which is composed of several one-hot-encoded properties. 

\subsection{Graph Construction Process}

Given a molecular graph defined as $G = (N, E)$ we propose to leverage the edge-specific information propagation framework described in \cite{ECMPNN} to learn a set of informative an embedding from which we can directly infer a graph. Our graph construction process is composed of two parts: 
\begin{itemize}
    \item An \textbf{Encoder} that in 
    \begin{itemize}
        \item \textbf{step 1} performs several spatial graph convolutions on the input graph, and in
        \item \textbf{step 2} aggregates those embeddings into a single graph latent representation.
    \end{itemize}
    \item A \textbf{Decoder} that in 
    \begin{itemize}
        \item \textbf{step 3} autoregressively  generates a set of continuous node embeddings conditioned on the learnt latent representation, and in
        \item \textbf{step 4} reconstructs the whole graph using edge-factorization.
    \end{itemize}
\end{itemize}

Figure \ref{fig:full_model} (a) and (b) provides a summary of those 4 steps.

\paragraph{Steps 1 and 2: Graph Representation Learning.} 
We use the Graph Convolutional Network (GCN) update rule \citep{gcnn} to encode the graph. Each node embedding can be written as a weighted sum of the edge-conditioned information of its neighbors in the graph. Namely for each $l$-th layer of the encoder, the representation is given by:
\begin{align}
H^l =  \sigma ( \sum_e [D_e^{-\frac{1}{2}}E_eD_e^{-\frac{1}{2}}H^{l-1}W_e^l ] + H^{l-1}W_s^l ) 
\end{align}
where $E_e$ is the $e$-th frontal slice of the adjacency tensor, $D_e$ the corresponding degree tensor and $W_e^l$ and $W_s^l$ are learned parameters of the layer.

Once we have the node embeddings we aggregate them to obtain a fixed-length latent representation of the graph. We propose to parametrize this aggregation step by an \textit{LSTM} and we compute the graph latent representation by a simple linear transformation of the last hidden state of this \textbf{Aggregator}:
\begin{align}
z = g_{agg}( f_{LSTM}^e (\{H^K\}) ).
\end{align}

Because the use of an LSTM makes aggregations permutation dependant, Like ~\cite{gsage}, we adapt the aggregator using randomly permuted sets of embeddings and empirically validated that this did not affect the performance of the model significantly.

In the subsequent steps we are interested in designing a graph decoding scheme from the latent code that is both scalable and powerful enough to model the interdependencies between the nodes and edges in the graph.

\paragraph{Step 3: Autoregressive Generation.}
We are interested in building a graph decoding scheme that models the nodes and their connectivity (represented by continuous embeddings $S$) in an autoregressive fashion. This is in contrast to \citep{martin, Molgan}, where each node and edge is conditionally independent given the latent code $z$. In practice this means that every detail of the interdependencies within the graph have to be encoded in the latent variable. 
We propose to tackle this drawback by autoregressive generation of the continuous embeddings $s = [s_0, s_1, ..., s_n]$ for $n$ nodes. More precisely we model the generation of node embeddings such that:
\begin{align}
p(s|z) = \prod_{i=1}^{n} p(s_{i}|s_{<i},z).
\end{align}

In our model, the autoregressive generation of embeddings is parametrized by a simple Long Short-Term Memory (LSTM, \cite{hochreiter1997long}) and is completely deterministic such that at each time step $t$ the LSTM decoder takes as input the previously generated embeddings and the latent code $z$ which captured node-invariant features of the graph. Each embedding is computed as a function of the concatenation of the current hidden state and the latent code $z$ such that:
\begin{align}
h_{t+1} = f_{LSTM}^d (g_{in}([z, s_{t}]), h_{t})\\
s_{t+1} = f_{embed}([h_{t+1}, z]) ,
\end{align}
where $f_{LSTM}^d$ corresponds to the LSTM recurrence operation and $g_{in}$ and $f_{embed}$ are parametrized as simple MLP layers to perform nonlinear feature extraction.

\paragraph{Step 4: Graph Decoding from Node Embeddings.}
As stated previously, we want to drive the generation of the continuous embeddings $s$ towards latent factors that contains enough information about the node they represent (i.e.  we can easily retrieve the one-hot atom type performing a linear transformation of the continuous embedding) and its neighbourhood (i.e. the adjacency tensor can be easily retrieved by comparing those embeddings in a pair-wise manner). For those reasons, we suggest to factorize each bond type in a  relational inference fashion \citep{decagon, NRI}. 

Let $S \in \mathbb{R}^{n \times  p}$ be the concatenated continuous node embeddings generated in the previous step. We reconstruct the adjacency tensor $E$ by learning edge-specific similarity measure for $k$-th edge type as follows:
\begin{align}
p(E_{:,:,k}|S) = \prod_{i=1}^{n}\prod_{j=1}^{n} p(E_{i,j,k}|s_{i},s_{j}).
\end{align}

This is modeled by a set of edge-specific factors $U = (u_{1}, \cdots, u_{e}) \in \mathbb{R}^{e \times  p}$ such that we can reconstruct the adjacency tensor as :
\begin{align}
\tilde{E}_{i,j,k} = \sigma (s_i^T D_k s_j) = p(E_{i,j,k}|s_i, s_j),
\end{align}
where $\sigma$ is the logistic sigmoid function, $D_{k}$ the corresponding diagonal matrix of the vector $u_{k}$ and the factors $(u_{i}) \in \mathbb{R}^{e \times p}$ are learned parameters.

We reconstruct the node features (i.e. the atom type) with a simple affine transformation such that:
\begin{align}
\tilde{N}_{i,:} = p(N_{i}|s_{i}) = \mbox{softmax}(W s_{i}),
\end{align}
where $W \in \mathbb{R}^{p \times d}$ is a learned parameter.


\paragraph{Generating Graphs of arbitrary sizes.} In order to generate graphs of different sizes we need to add what we call here an \textbf{Existence} module that retrieves a probability of a node belonging to the final graph for each of the embedding generated (in step 3). This module is parametrized as a simple MLP layer followed by a sigmoid activation and stops the unrolling of the embedding LSTM generator whenever we encounter a \textit{non-informative} embedding. This module can be interpreted as an $<eos>$ translator.

\subsection{Training}
\paragraph{Teacher forcing.}
To make the model converge in reasonable time we adapt teacher-forcing on language models~\citep{TF} as follows. The training is thus done in 3 phases:
\begin{itemize}
  \item We first pre-train the GCN part along with the embedding decoder (factorization, nodes and existence modules) to reconstruct the graphs. This corresponds to the training of a simple Graph AE as in \cite{VAE_kipf} except that we also want to reconstruct the nodes' one-hot features (and not just the \textit{relations}).
  \item We then append those two units to the embedding aggregator and generator while keeping them fixed. In this second phase, the embedding generator is trained using teacher forcing where at each time step $t$ the \textit{LSTM} decoder does not take as input the previously generated embedding but the \textit{true} one that is the direct output of the pretrained GCN embedding encoder.
  \item Finally in order to transition from teacher-forcing to a fully autoregressive state we increasingly \citep{SS} feed the LSTM generator more of its own predictions. When a fully autoregressive state is reached the pre-trained units are unfrozen and the whole model continues training end-to-end.
\end{itemize}

\paragraph{Log-Likelihood Estimates} We train the autoencoder on the reconstruction error using the MLE with the estimate negative log-likelihood  given by:
\begin{gather}
\mathcal{L}_{rec} = \mathcal{L}_{X} + \mathcal{L}_{\bar{X}} + \mathcal{L}_{N}
\end{gather}
where $n$ the number of nodes in the graph, $X$ and $\bar{X}$ corresponding to the existing and non existing edges in the adjacency tensor $E$, and $N$ is the node features, such that:
\begin{gather}
\mathcal{L}_{X} =-\frac{1}{|X|}\sum_{(i,j) \in X} E_{i,j,:}^T \log(\tilde{E}_{i,j,:}) + (1 -E_{i,j,:})^T \log(1 -\tilde{E}_{i,j,:})\\
\mathcal{L}_{\bar{X}} =-\frac{1}{|\bar{X}|}\sum_{(i,j) \in \bar{X}} \sum_{k} \log(1 -\tilde{E}_{i,j,k}) \\
\mathcal{L}_{N} =-\frac{1}{n} \sum N^T\log(\tilde{N}),
\end{gather}
Since molecular graphs are sparse,  we found that such separate normalisations were helpful for the training.

\subsection{Conditional Generation and Optimisation}
\paragraph{Model overview.}

Given the entangled latent code $z$ for a given input molecular graph, we create a conditioned input $(z, y)$ by augmenting $z$ with a set of structured attributes $y$ - the target properties of interest, such as physico-chemical property. The conditional generator is then trained on the combined reconstruction and property loss.  At the end of a successful training we expect the decoder to generate samples that have the properties specified in $y$ and to be \textit{similar} (in terms of information contained in $z$) to the original query molecular graph (encoded as $z$). To do so we choose a mutual information maximization approach (detailed in the Appendix~\ref{ap:mutual-info}) that involves the use of discriminators that assess the properties $\tilde{y}$ of the generated samples and their feedback is used to guide the learning of the generator.


\paragraph{Discriminator Pre-Training}

In this phase we pre-train a discriminator to assess the property $y$ of a generated sample so that we can backpropagate its feedback to the generator (the discriminator can be trained on another dataset and we can have several discriminators for several attributes of interest).
In order to have informative gradients in the early stages of the training we have trained the discriminator on continuous approximations of the discrete training graphs (details in Appendix~\ref{ap:conditional-setting}) so that our objective becomes:
\begin{align}
    \mathcal{L}_{dis} =\mathbb{E}_{(x,y) \sim \tilde{p}_{data}(x,y)}[-\log Q(y|x)],
\end{align}
where the graphs sampled from $\tilde{p}_{data}(x)$ are the probabilistic approximations of the discrete graphs from the training distribution $p_{data}(x)$.

The next step is to incorporate the feedback signal of the trained discriminator in order to formulate the property attribute constraint. The training is decomposed in two phases in which we learn to reconstruct graphs of the dataset (\textbf{MLE} phase) and to modify chemical attributes (\textbf{Variational MI maximization} phase).

\paragraph{Encoder Learning.}
The encoder is updated only during the reconstruction phase where we sample attributes $y$ from the true posterior. The encoder loss is a linear combination of the molecular graph reconstruction ($\mathcal{L}_{rec}$) and the property reconstruction ($\mathcal{L}_{prop}$). The total encoder loss is:
\begin{align}
    \mathcal{L}_{enc} =  \mathcal{L}_{rec} + \beta \mathcal{L}_{prop}.
\end{align}
where $\mathcal{L}_{rec} = \mathbb{E}_{(x,y) \sim p_{data}(x,y), z \sim E(z|x)}[-\log p_{gen}(x|z,y)]$  (using the log-likelihood estimates in \textbf{(7)}) and $\mathcal{L}_{prop} =  \mathbb{E}_{(x,y) \sim p_{data}(x,y), z \sim E(z|x), x'\sim p_{gen}(x|z,y) }[-\log Q(y|x')]$.
With $\beta \in [0,1]$ a hyperparameter of the model.


\paragraph{Generator Learning.}
The generator is updated in both reconstruction and conditional phases. In the \textbf{MLE} phase the generator is trained with same loss $\mathcal{L}_{enc}$ as the encoder so that it is pushed towards generating realistic molecular graphs. In the \textbf{MI maximization} phase we sample the attributes from a prior $p(y)$ s.t. we minimize the following objective: $\mathcal{L}_{cond} =\mathbb{E}_{x \sim p_{data}(x),y \sim p(y) z \sim E(z|x), x'\sim p_{gen}(x|z,y)}[-\log  Q(y|x')] $,
\begin{align}
\mathcal{L}_{gen} = \mathcal{L}_{rec} + \alpha \mathcal{L}_{cond} + \beta \mathcal{L}_{prop}.
\end{align}
where $\beta, \alpha \in [0,1]$ are hyperparameters of the model.

In this phase the only optimisation signal comes from the trained discriminator. Since there are no \textit{realism} constraint specified in our model (see ~\cite{JTVAE, Lesko}), there is a risk of ``falling off the manifold''. A possible way solution mitigate against it is to add a similarity discriminator trained to distinguish between the real probabilistic graph and the generated ones - so that when trying to satisfy the attribute constraint the generator is forced to produce valid molecular graphs. We leave this for future work. 


\begin{table}[t!]
\vspace*{2mm}
\label{rec_table}
\begin{center}
\begin{tabular}{ll}
\multicolumn{1}{l}{\bf Method}  &\multicolumn{1}{l}{\bf Reconstruction Accuracy} \\
 \toprule
JT-VAE~\cite{JTVAE}  &76.7 \\
\midrule 
JT-AE (without stereochemistry) & 69.9\\
\midrule 
DEFactor - 56 & 89.2 \\
DEFactor - 64 & 89.4 \\
\textbf{DEFactor - 100} &\textbf{89.8}\\
\bottomrule
\end{tabular}
\caption{Molecular graph reconstruction task. We compare the performance of our decoder in the molecular reconstruction task with the JT-VAE. The results for JT-VAE result is taken from~\cite{JTVAE} and use a latent code of size 56. The JT-AE refers to an adapted version of the original model using the same parameters. It is however deterministic, and like DEFactor does not evaluate stereochemistry.}
~\label{table:reconstruction_acc}
\end{center}
\end{table}

\section{Experiments}\label{sec:experiments}
To compare with recent results in constrained molecular graph optimization~\cite{JTVAE, Lesko}, we present the following experiments : 
\begin{itemize}
    \item \textbf{Molecular Graph Reconstruction}: We test the autoencoder framework on the task of reconstructing input molecules from their latent representations. 
    \item \textbf{Conditional Generation}: We test our conditional generation framework on the task of generating novel molecules that satisfy a given input property. Here, we are interested in the octanol-water partition coefficient (LogP) optimization used as benchmark in~\cite{pmlr-v70-kusner17a, JTVAE, Lesko}.
    \item \textbf{Constrained Property Optimization}: Finally, we test our conditional autoencoder on the task of modifying a given molecule to improve a specified property, while constraining the degree of deviation from the original molecule. Again we use the LogP benchmark for the experiment.
\end{itemize}

Finally in the following section  we use the 250K subset of the ZINC~\cite{zinc} dataset, released by~\cite{pmlr-v70-kusner17a}, along with their given train and test splits. 

\paragraph{Molecular graph reconstruction:} In this task we evaluate the exact reconstruction error from encoding and decoding a given molecular graph from the test set. We report in Table~\ref{table:reconstruction_acc} the ratio of exactly reconstructed graphs, where we see that the our autoencoder outperforms the JT-VAE~\cite{JTVAE} which has the current state-of-the-art performance in this task. Appendix~\ref{ap:reconstruction-number-atoms} reports the reconstruction ratio as a function of the molecule size (number of heavy atoms).


\begin{figure}[h]
\centering
\includegraphics[scale =0.6]{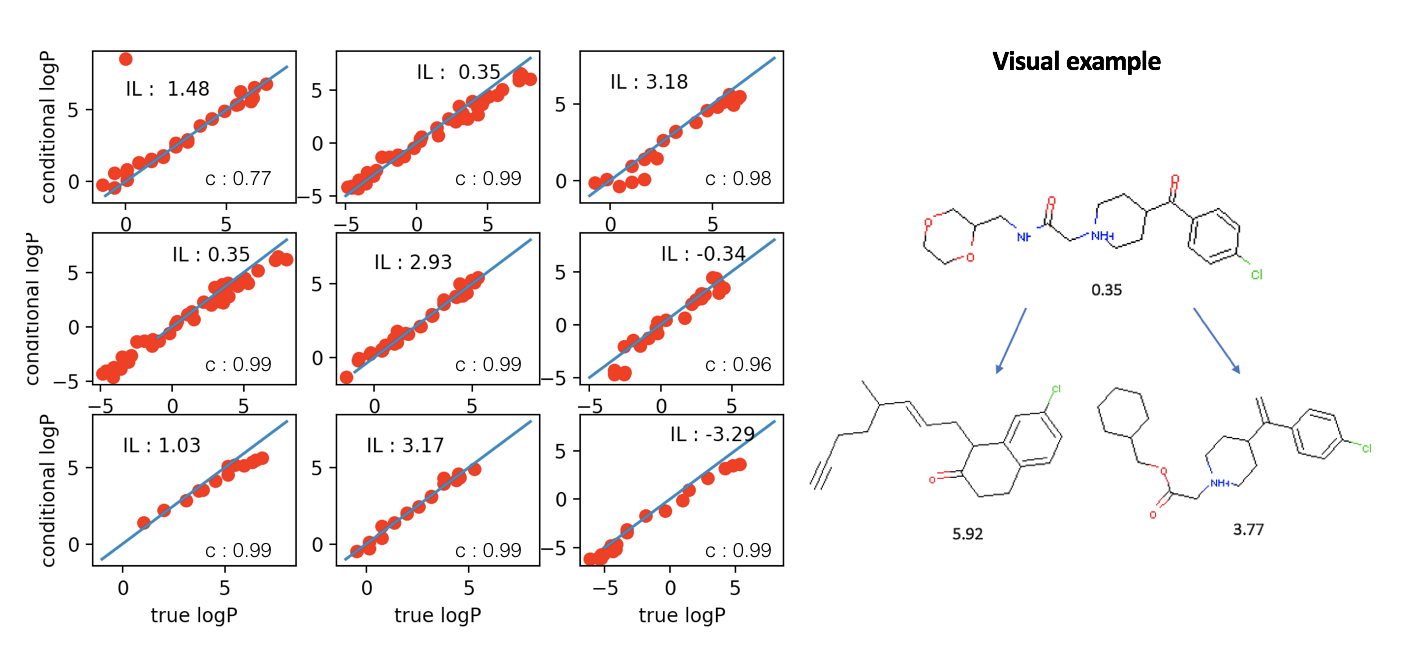}
\caption{Conditional generation: The initial LogP value of the query molecule is specified as \textit{IL} and the Pearson correlation coefficient is specified as $c$. We report on the y-axis the conditional value given as input and on the x-axis the true LogP of the generated graph when translated back into molecule. For each molecule we sample uniformly around the observed LogP value and report the LogP values for the  decoded graphs corresponding to valid molecules.}
\label{fig:scatter}
\end{figure}

\paragraph{Conditional Generation:} In this task we evaluate the conditional generation formulation described in Section 3.3. For a given molecule $m$ with an observed property value $y$, the goal here is to modify the molecule to generate a new molecule with the given target property value; ($m^{*}, y^{*}$). 
New molecules are generated by conditioning the decoder on $(z; y^{*})$, where $z$ is the latent code for $m$. The decoded new molecule $m^{*}$, is ideally best suited to satisfy the target property. This is evaluated by comparing the property value of the new molecule with the target property value.  A generator that performs well at this task will produce predicted molecules with property values that are close to the target. In these experiments, LogP was chosen as the desired property, and we use RDKIT~\cite{rdkit} to calculate the LogP values of generated molecules.



The scatter plots in Figure~\ref{fig:scatter} give for a  selected set of test molecules, the correlation of target property values against the evaluated property value of the correctly decoded molecules. Here, for each molecule the target set is defined by uniformly sampling around the  observed property value for the molecule.

\paragraph{Constrained Property Optimization:} In this section we follow the evaluation methodology outlined in~\cite{JTVAE, Lesko}, and evaluate our model in a constrained molecule property optimization. In contrast to~\cite{JTVAE}, due to the conditional formulation, does not need retraining for the optimisation task. 

Given the 800 molecules with the lowest penalized LogP\footnote{The penalized logP is octanol-water partition coefficient (logP) penalized by the synthetic accessibility (SA) score and the number of long cycles, see~\cite{JTVAE}} property scores from the test set, we evaluate the decoder by providing pairs of ($z, y^{*}$) with increasing property scores, and among the valid decoded graphs we compute:
\begin{itemize}
    \item Their similarity scores (Sim.) to the encoded target molecule (called the prototype); 
    \item Their penalized LogP scores. Note that in this setting the conditioning property values ($y^{*}$) are the unpenalized LogP scores. However, to evaluate the model we compute the penalized LogP scores to assess the model's ability to decode synthetically accessible molecules.
    \item While varying the similarity threshold values ($\delta$), we compute the success rate (Suc.) for all 800 molecules. This measures how often we to get a novel molecule with an improved penalized LogP score.
    \item Finally, for different similarity thresholds, for successfully decoded molecules, we report the average improvements (Imp.) and the similarity (Sim.) for the molecule that is most improved. We compare our results with \cite{JTVAE, Lesko}.
\end{itemize}

The final results are reported in Table~\ref{table:constrained-op}. As can be seen, although slightly behind GCPN~\cite{Lesko} w.r.t. success rates (Suc.), DEFactor significantly outperforms other models in terms of improvements (Imp.) achieved (by between $1.3\times$ and $1.95\times$ for thresholds 0.2 and 0.6 respectively, with respect to the next best model GCPN).

\begin{table}[tb]
    \centering
    \scriptsize
  
    \renewrobustcmd{\bfseries}{\fontseries{b}\selectfont}
    \renewrobustcmd{\boldmath}{}
    \newrobustcmd{\B}{\bfseries}
  
  \begin{tabular}{cccccccccc}
    \toprule
    \multirow{2}{*}{$\delta$} & \multicolumn{3}{c}{JT-VAE} & \multicolumn{3}{c}{GCPN} & \multicolumn{3}{c}{DEFactor} \\
    \cline{2-10}
    & Imp. & Sim. & Suc. & Imp. & Sim. & Suc. & Imp. & Sim.& Suc. \\
    \midrule
    \B 0.0 & 1.91$\pm$ 2.04 & 0.28$\pm$0.15 & 97.5$\%$ & 4.20$\pm$1.28 & \B 0.32$\pm$0.12 & \B 100$\%$ & \B 6.62$\pm$2.50 & 0.20$\pm$0.16 & 91.5$\%$ \\
    \B 0.2 & 1.68$\pm$ 1.85 & 0.33$\pm$0.13 & 97.1$\%$ & 4.12$\pm$1.19 & \B 0.34$\pm$0.11 & \B 100$\%$ & \B 5.55$\pm$2.31 & 0.31$\pm$0.12 & 90.8$\%$ \\
    \B 0.4 &  0.84$\pm$ 1.45 & \B 0.51$\pm$0.10 & 83.6$\%$ & 2.49$\pm$1.30 & 0.47$\pm$0.08 & \B 100$\%$ & \B 3.41$\pm$1.8 & 0.49$\pm$0.09 & 85.9$\%$ \\
    \B 0.6 &  0.21$\pm$ 0.71 & 0.69$\pm$0.06 & 46.4$\%$ & 0.79$\pm$0.63 & 0.68$\pm$0.08 & \B 100$\%$ & \B 1.55$\pm$1.19 & \B 0.69$\pm$0.06 & 72.6$\%$ \\
    \bottomrule
    
  \end{tabular}
  
  \caption{Constrained penalized LogP maximisation task: each row gives a different threshold similarity constraint $\delta$ and columns are for improvements (Imp.), similarity to the original query (Sim.), and the success rate (Suc.). Values for other models are taken from \cite{Lesko}.}
  \label{table:constrained-op}
\end{table}

\section{Future work}\par
In this paper, we have presented a new way of modelling and generating graphs in a conditional optimisation setting such that the final graph being fully differentiable w.r.t to the model parameters. We believe that our \textit{DEFactor} model will contribute to understanding and building ML-driven applications for de-novo drug design or generation of molecules with optimal properties, without resorting to methods that do not directly optimise the desired properties.

Note that a drawback of our model is that it uses an  MLE training process which forces us to either fix the ordering of nodes or to perform a computationally expensive graph matching operation to compute the loss. Moreover in our fully deterministic conditional formulation we assume that chemical properties optimisation is a one-to-one mapping but in reality there may exist many suitable way of optimizing a molecule to satisfy one property condition while staying similar to the query molecule. To that extent it could be interesting to augment our model to include the possibility of a one-to-many mapping. Another way of improving the model could also be to include a validity constraint formulated as training a discriminator that discriminates between valid and generated graphs.

\section*{Acknowledgements}
The authors thank NSERC and CIFAR for funding at U. Montreal and Mila as well as Marco Fiscato, Petar Veličković, Jian Tang and Benedek Fabian for useful discussions.

\bibliography{main}
\bibliographystyle{unsrt}

\newpage
\begin{appendices}
\section{}

\subsection{Models Comparison}

\newcommand{\cmark}{\ding{51}}%
\newcommand{\xmark}{\ding{55}}%
\begin{figure}[h]
    \centering
  \begin{tabular}{c||ccccc}
    \textbf{Model} & Inference  & Parameters & Constrained& Probabilistic & No Retraining \\
    \midrule
   
    MolGAN \cite{Molgan} & \xmark & \xmark& \xmark& \cmark& NA \\
   JT-VAE \cite{JTVAE} & \cmark &\cmark&  \cmark& \xmark& \xmark \\
    GCPNN\cite{Lesko} & \xmark &\cmark&  \cmark& \xmark& \cmark \\
    \midrule
    DEFactor(Ours) & \cmark & \cmark& \cmark& \cmark& \cmark 

\end{tabular}
\caption{We report here a comparison of the abilities of previous recent models involving molecular graph generation and optimization.}
\end{figure}

We are interested in the following features of the models : 
\begin{itemize}
    \item \textbf{Inference} : If the model is equipped or not with an inference network. To encode some target molecule like we do in the conditional setting.
     \item \textbf{Parameter-efficient} : If the number of parameters of the model depends on the graph sizes.
      \item \textbf{Constrained} : If the model is studied in a constrained optimization scenario : namely the case where we want to optimize a property while  constraining the degree of deviation from the original molecule.
       \item \textbf{Probabilistic} : If the outptut of the model is a probabilistic graph s.t. it is differentiable w.r.t to the decoder's parameters.
        \item \textbf{No Retraining} : If we need to retrain/fine-tune/perform gradient-ascent each time we want to optimize a novel molecule.
\end{itemize}

\section{}

\subsection{Graphs continuous approximation}\label{ap:conditional-setting}
For the pre-training of the discriminators we suggested to train them on continuous approximation of the discrete graphs that \textit{resembles} the output of the decoder. To that extent we used the trained partial graph autoencoder (used for the teacher forcing at the beginning of the training of the full autoencoder.)
 \begin{figure}[h]
\centering
\includegraphics[scale = 0.7]{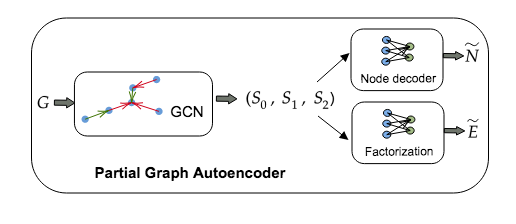}
\label{embedae}
\caption{Partial graph Autoencoder used for the pre-training.}
\end{figure}

\subsection{Mutual information maximization}\label{ap:mutual-info}

For the conditional setting we choose a simple mutual information maximization formulation. The objective is to maximize the MI $I(X;Y)$ between the target property $Y$ and the decoder's output $X = G_{\theta}(Y)$ under the joint $p_{\theta}(X,Y)$ defined by the decoder $G_{\theta}$. In the conditional setting $G_{\theta}$ is also conditioned on the encoded molecule $z$ but for simplicity we treat it as a parameter of the decoder (and thus reason with one target molecule from which we want to modify attributes). We define the MI as:


\begin{align*}
I(y;G_{\theta}(y)) &= \mathbb{E}_{x \sim G_{\theta}(y)}[\mathbb{E}_{y\prime \sim p_{\theta}(y|x)}[\log p_{\theta}(y\prime|x)]]+ H(y) \\
&= \mathbb{E}_{x \sim G_{\theta}(y)}[D_{KL}(p_{\theta}(.|x)||Q(.|x)) \\&+\mathbb{E}_{y\prime \sim p_{\theta}(y|x)}[\log Q(y\prime|x)]] + H(y) \\
&\geq \mathbb{E}_{x \sim G_{\theta}(y)}[\mathbb{E}_{y\prime \sim p_{\theta}(y|x)}[\log Q(y\prime|x)]] + H(y)
\end{align*}

In our conditional setting we pre-trained the discriminators (parametrized by $Q$ in the lower bound derivation) to approximate $p_{data}(y|x)$ which makes the bound tight only when $p_{\theta}(y_{paired}|x)$ is close to $p_{data}(y|x)$ and this corresponds to a stage where the decoder has maximized the log-likelihood of the data well enough (i.e. when it is able to reconstruct input graphs properly when $z$ and $y$ are paired).
Thus, in the conditional setting we are maximizing the following objective: 

\begin{align*}
\mathcal{L}_{cond} = \mathbb{E}_{x,y \sim p_{data}(x,y) , z \sim E(x), y'\sim p(y) } [ \log G_{\theta}(y,z) +  I(y\prime; G_{\theta}(y\prime,z))] 
\end{align*}

\section{}
\subsection{Reconstruction as a function of number of atoms}~\label{ap:reconstruction-number-atoms}
\begin{figure}[h]
\centering
\includegraphics[scale = 0.4]{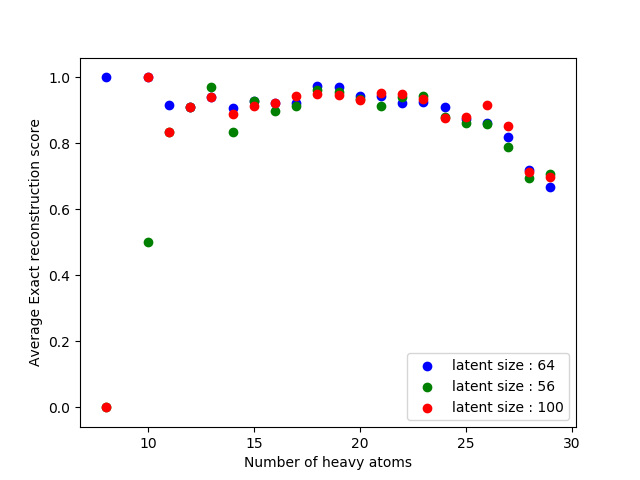}
\label{embedae}
\caption{Accuracy score  as a function of the number of heavy atoms in the molecule(x axis) for different size of the latent code.}
\end{figure}

Notice that as we make use of a simple LSTM to encode a graph representation, there is a risk that for the largest molecules the long term dependencies of the embeddings are not captured well resulting in a bad reconstruction error. We capture this observation in figure 4. One possible amelioration could be to add at each step another non-sequential aggregation of the embeddings (average pooling of the embeddings for example) or to make the encoder more powerful by adding some attention mechanisms. We leave these options for future work.
\newpage
\subsection{Visual similarity samples}\label{appendix:vis-sim-samples}
\begin{figure}[h]
\centering
\includegraphics[scale = 0.53]{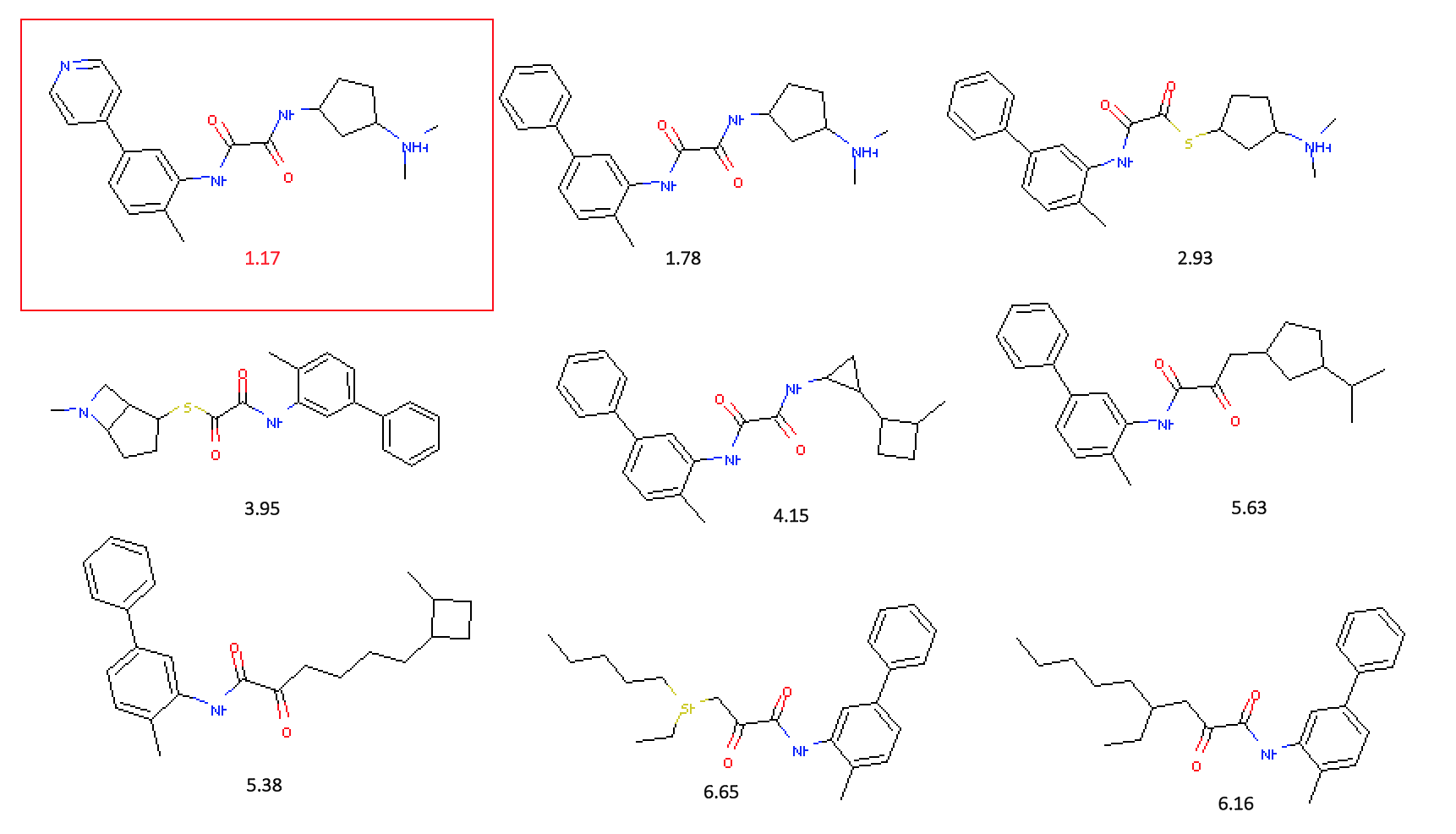}
\vspace{1cm}
\includegraphics[scale = 0.53]{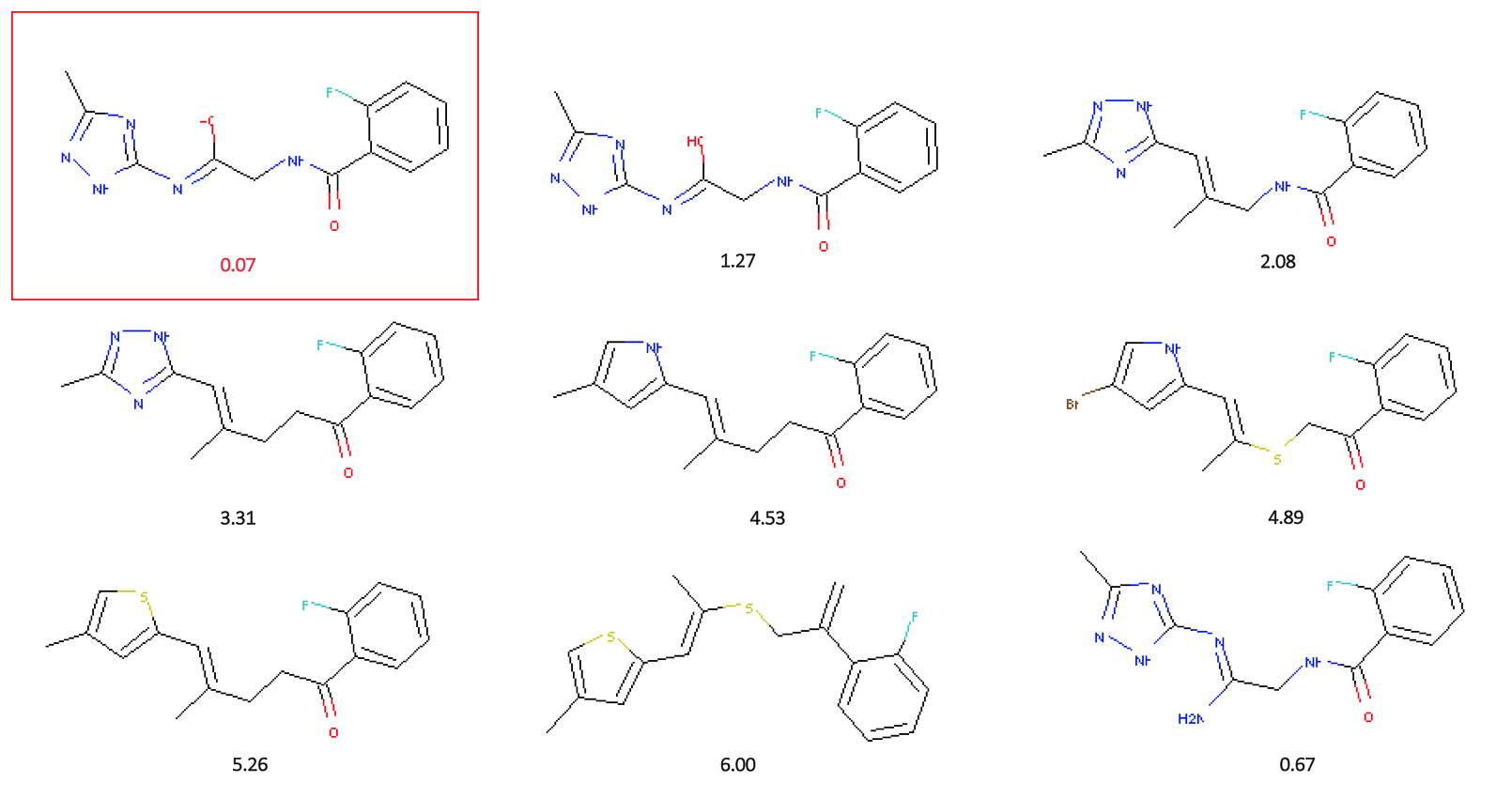}
\end{figure}
\begin{figure}[t!]
\includegraphics[scale = 0.54]{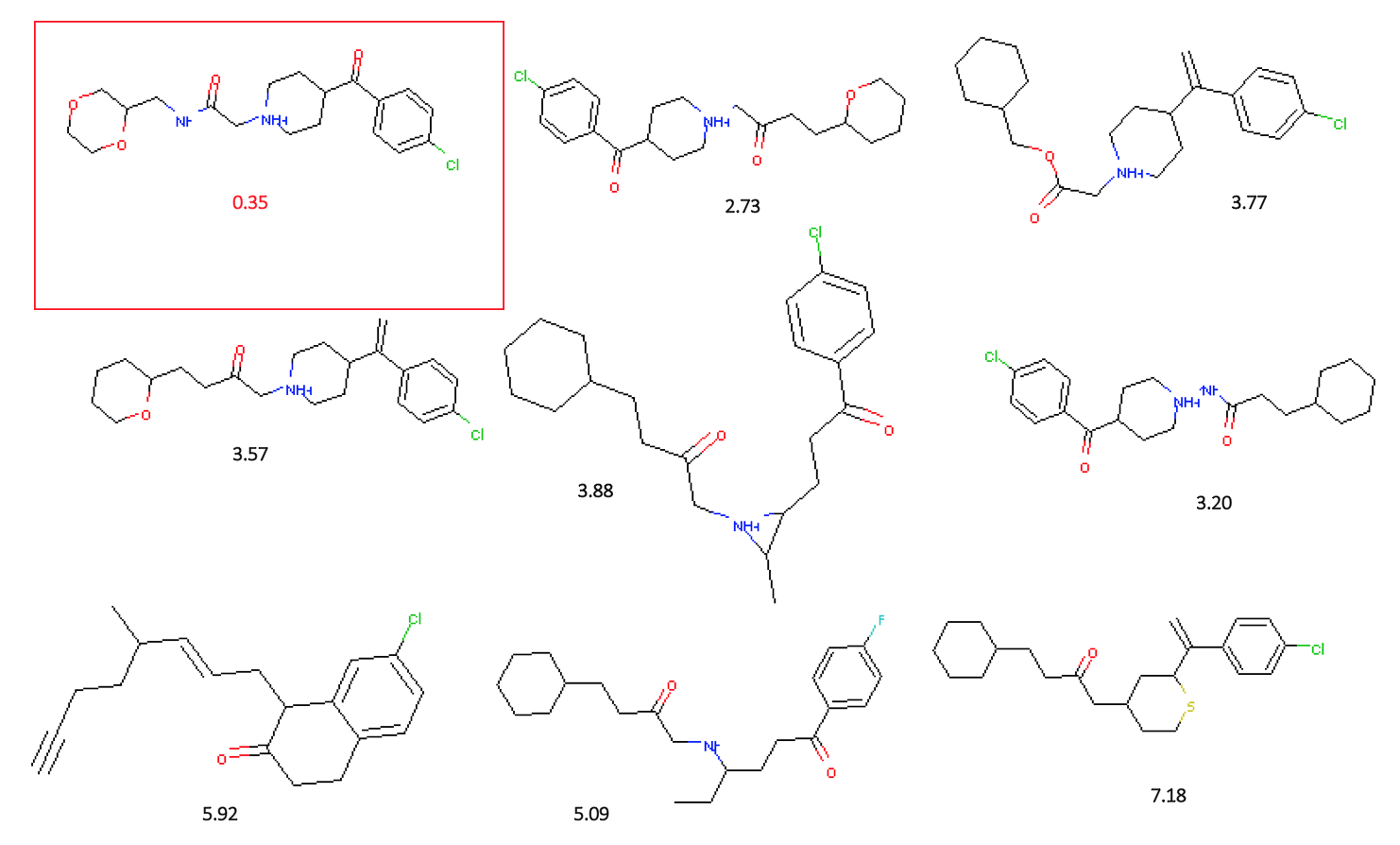}
\label{27}
\caption{LogP increasing task visual example. The original molecule is outlined in red.}
\end{figure}


\end{appendices} 

\end{document}